\documentclass[runningheads]{llncs}
\usepackage{graphicx}
\usepackage{amsmath,amssymb} 
\usepackage{color,epstopdf,subcaption}
\begin{document}

\newcommand{\point}{
    \raise0.7ex\hbox{.}
    }


\pagestyle{headings}

\mainmatter

\title{Distinguishing Posed and Spontaneous Smiles by Facial Dynamics} 

\titlerunning{Distinguishing Posed and Spontaneous Smiles by Facial Dynamics} 

\authorrunning{Mandal \emph{et al.}} 

\author{Bappaditya Mandal$^{*,1}$, David Lee$^2$ and Nizar Ouarti$^3$}
\institute{$^1$Visual Computing Department, Institute for Infocomm Research, Singapore,\\$^2$Electrical and Computer Engineering, National University of Singapore, \\$^3$Université Pierre et Marie Curie, Sorbonne Universités, France.\\ Email address: bmandal@i2r.a-star.edu.sg ($^*$Contact author: Bappaditya Mandal); a0097074@u.nus.edu (David Lee); nizar.ouarti@ipal.cnrs.fr (Nizar Ouarti)}

\maketitle

\begin{abstract}
Smile is one of the key elements in identifying emotions and present state of mind of an individual. In this work, we propose a cluster of approaches to classify posed and spontaneous smiles using deep convolutional neural network (CNN) face features, local phase quantization (LPQ), dense optical flow and histogram of gradient (HOG). Eulerian Video Magnification (EVM) is used for micro-expression smile amplification along with three normalization procedures for distinguishing posed and spontaneous smiles. Although the deep CNN face model is trained with large number of face images, HOG features outperforms this model for overall face smile classification task. Using EVM to amplify micro-expressions did not have a significant impact on classification accuracy, while the normalizing facial features improved classification accuracy. Unlike many manual or semi-automatic methodologies, our approach aims to automatically classify all smiles into either `spontaneous' or `posed' categories, by using support vector machines (SVM). Experimental results on large UvA-NEMO smile database show promising results as compared to other relevant methods.
\end{abstract}

\section{Introduction}

In the past, the research on human affect was focused on basic emotions which include happiness, sadness, fear, disgust, anger, and surprise \cite{Zeng1}. Past research conducted in this area largely focused on posed facial expressions. Primarily, because of difficulty in obtaining and capturing spontaneous facial expression that lead to unavailability of such databases.  However, these discrete emotions (posed facial expressions) fail to describe the wide range of emotions that occur in natural social interactions. Recent studies have shown that spontaneous facial expressions reveal more information about the emotions of the person, and spontaneous expressions differ greatly from its posed counterpart \cite{Ekman5}. For example, a spontaneous smile is generally interpreted to show enjoyment or happiness, but it can also arise out of frustration \cite{Hoque1}. As a result, more emphasis has been placed on spontaneous facial expressions in the recent years. Smile is one of the key elements in identifying emotions and state of mind from facial expressions, as it is frequently exhibited to convey emotions like amusement, politeness and embarrassment \cite{Ambadar1}, and is also used to mask other emotional expressions \cite{Ekman3}. Since smile is the easiest expression to pose \cite{Ekman3}, it is important for the machines (and also humans) to distinguish when it is posed and when it is a genuine smile of enjoyment.

This work contributes to the field of affective computing by solving the problem of distinguishing posed and spontaneous smiles. Development in this field can be applied commercially to enhance our daily lives. For example, automatic human affect recognition can be applied to wearable devices like Google Glass to help children with autism who have difficulty reading expressions to understand emotions in a social environment \cite{Hadwin1,Xu2,Mandal13,Mandal14}. It can also be installed in vehicles to detect fatigue levels of the driver and prevent fatigue-causing accidents from occurring. Affective computing can also be applied in the academic fields like psychology, psychiatry, behavioral science and neuroscience, to reduce the time-consuming task of labeling human affects and can improve human lives.

\subsection{Related Work}
In early years of research, it was thought that the morphological features of the face were good indicators of a spontaneous smile. The Facial Action Coding System (FACS) \cite{Ekman4} defines Action Units (AU), which are the contraction or relaxation of one or more muscles. It is commonly used to code facial expressions, and can be used to identify emotions. In FACS, a smile corresponds to AU12, which is the contraction of the zygomatic major muscle that raises the lip corners. A genuine smile of joy is thought to include AU6, also known as the Duchenne Marker, which is the contraction of the orbicularis oculi (pars lateralis) muscle that raises the cheek, narrows the eye aperture, and forms wrinkles on the external sides of the eyes. However, recent research casts doubt on the reliability of Duchenne marker in identifying true feelings of enjoyment \cite{Krumhuber1}. Another possible marker of spontaneous smiles is the symmetry of the smile. Initial studies claim that smile symmetry is a factor in identifying spontaneous smiles, where spontaneous smiles are more symmetrical than posed smiles \cite{Ekman1}. However, later studies report no significant differences of symmetry \cite{Dibeklioglu1,Schmidt1}.

Recently, more attention has been paid to dynamical properties of smiles such as the duration, amplitude, speed, and acceleration instead of static features like smile symmetry or the AUs. To analyze these properties, the smile is generally broken up into three different phases - onset, apex, and offset. Spontaneous smiles tend to have a smaller amplitude, a slower onset \cite{Cohn1}, and a shorter total duration \cite{Schmidt2}. The eye region is analyzed as well - the eyebrow raise in posed smiles have a higher maximum speed, larger amplitude and shorter duration than spontaneous ones \cite{Schmidt1}. Most techniques extract dynamic properties of smiles that are known to be important factors in classifying smiles \cite{Zeng1,Valstar2}. Apart from these properties of smiles, facial dynamics can reveal other useful information for the classification of smiles, such as the subject's age.

Currently, the method with the best performance is the one proposed by Dibeklioglu \textit{et al.} \cite{Dibeklioglu1}. In their method, 11 facial feature points are tracked using a Piecewise Bezier Volume Deformation (PBVD) tracker, which was proposed by Tao and Huang \cite{Tao1}. The duration, amplitude, speed and acceleration of various features in the eyes, cheeks and mouth regions are calculated across the 3 different phases of the smile, and a mid-level fusion is used to concatenate the features. Using these features, the classification accuracy on the UvA-NEMO Smile Database was 87.0\%. An optical flow along with various face component based features is proposed in \cite{Mandal11}, where it is shown that even optical based features can perform similar to face component based features. However, their tracking is initialized by manually annotated facial landmarks.

This work aims to discover other additional features from facial dynamics that can be useful for classification. To do so, the entire face region should be processed and used for smile classification instead of extracting the known features of a smile described above. A cluster of approaches are developed to extract features from a video and use a classifier to determine whether the smile in the video is posed or spontaneous. The system first pre-processes the video by tracking the face and extracting the face region from the video after normalization. 3 different normalization techniques are tested to determine its efficacy. Micro-expressions in the face region are amplified to test their impact on smile classification. Then, several image processing techniques are used to extract features from the face region. The features are then post-processed to reduce the dimensionality of the features and to normalize the number of features per video for the classifier to work. Finally, the processed features are given as an input to a support vector machines (SVM) to classify between posed and spontaneous smiles.

In the rest of this paper, in section 2, a cluster of methodologies is proposed to test the effectiveness of three normalization techniques, and the effectiveness of HOG, LPQ, dense optical flow and pre-trained convolutional neural network features for smile classification. A relatively new technique for amplifying micro-expressions is also tested to analyze the impact of micro-expressions in classifying smiles. In section 3, experimental results and analysis are presented. Conclusions are drawn in section 4.

\section{Proposed Methodologies}
\begin{figure}[!ht]
\begin{center}
\scalebox{0.55}{\rotatebox{0}{\includegraphics*{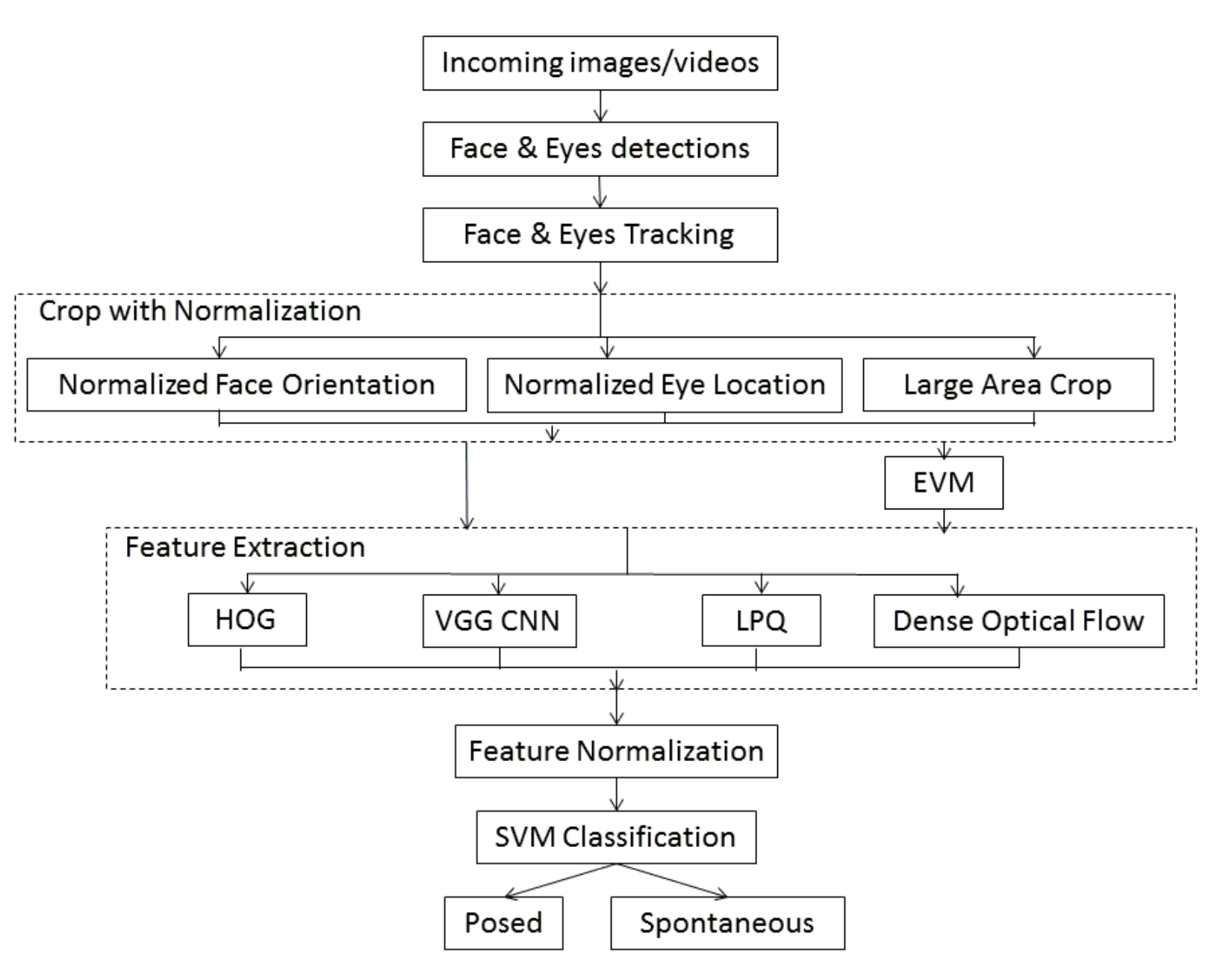}}}
\caption{Block diagram of the proposed system.}\label{sysDiagram}
\end{center}
\end{figure}

Fig. \ref{sysDiagram} shows the flow of our proposed approaches.
The incoming videos/images are pre-processed to extract the face and eye regions from the image frames. Faces are then tracked using these detected regions so as to understand the global movement of the faces. 3 separate normalization techniques are used to normalize the face in the extracted image sequence, and each technique is tested to see how effective each one is. Eulerian video magnification (EVM) is used in the pre-processing step to amplify micro-expressions in the face region to test the impact of micro-expressions in classification. After the face region is extracted, the videos are processed to extract features for classification. Different feature extraction techniques are used to test the effectiveness of each method. The features are then post-processed to reduce the dimensionality of the features and to normalize the number of features per video for the classifier to work. Finally, the processed features are given as an input to the SVM to classify between posed and spontaneous smiles. Each of these blocks are discussed below.

\subsection{Pre-processing Techniques}
The incoming image frames are pre-processed using three different methodologies which are described below:

\subsubsection{Face \& Eye Detections:}
The face and eyes are first detected to be used for tracking in the next step. Initially, a cascade object detector from Matlab's computer vision toolbox was implemented to detect the face and both eyes in the first frame of the video. However despite fine-tuning, it was inaccurate as there were many cases of false positives. Such errors will propagate throughout the entire system and cause anomalous results in the classification. Therefore this implementation was discarded and replaced by a more accurate method by Mandal \emph{et al.} \cite{Mandal5,Mandal12}. This method uses a fusion of OpenCV face and eye detectors \cite{OpenCV1} and Integration of Sketch and Graph patterns (ISG) eye detectors developed for human-robot-interaction by Yu \emph{et al.} in \cite{Yu2}. Through the integration of both eye detectors \cite{Mandal10,Chakraborty1}, we are able to achieve high success rate of 96.3\% eye localization in the smile face images for both frontal and semi-frontal faces at various scales with large global motions \cite{Mandal9}. The output of this step is 3 bounding boxes containing the face, the left eye and the right eye, separately.

\subsubsection{Face \& Eye Tracking:}
\begin{figure}[!ht]
\begin{center}
\scalebox{0.258}{\rotatebox{0}{\includegraphics*{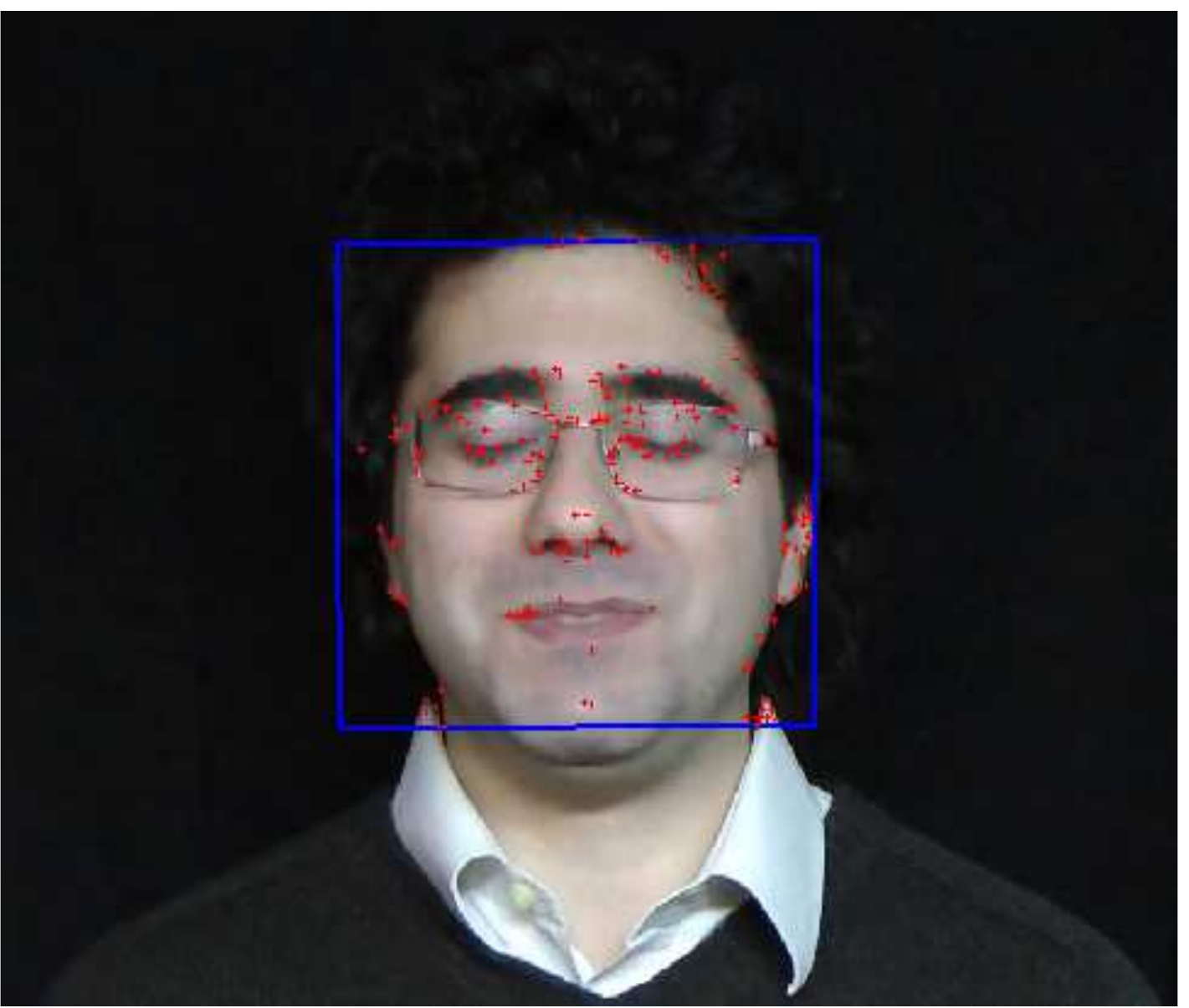}}}
\scalebox{0.25}{\rotatebox{0}{\includegraphics*{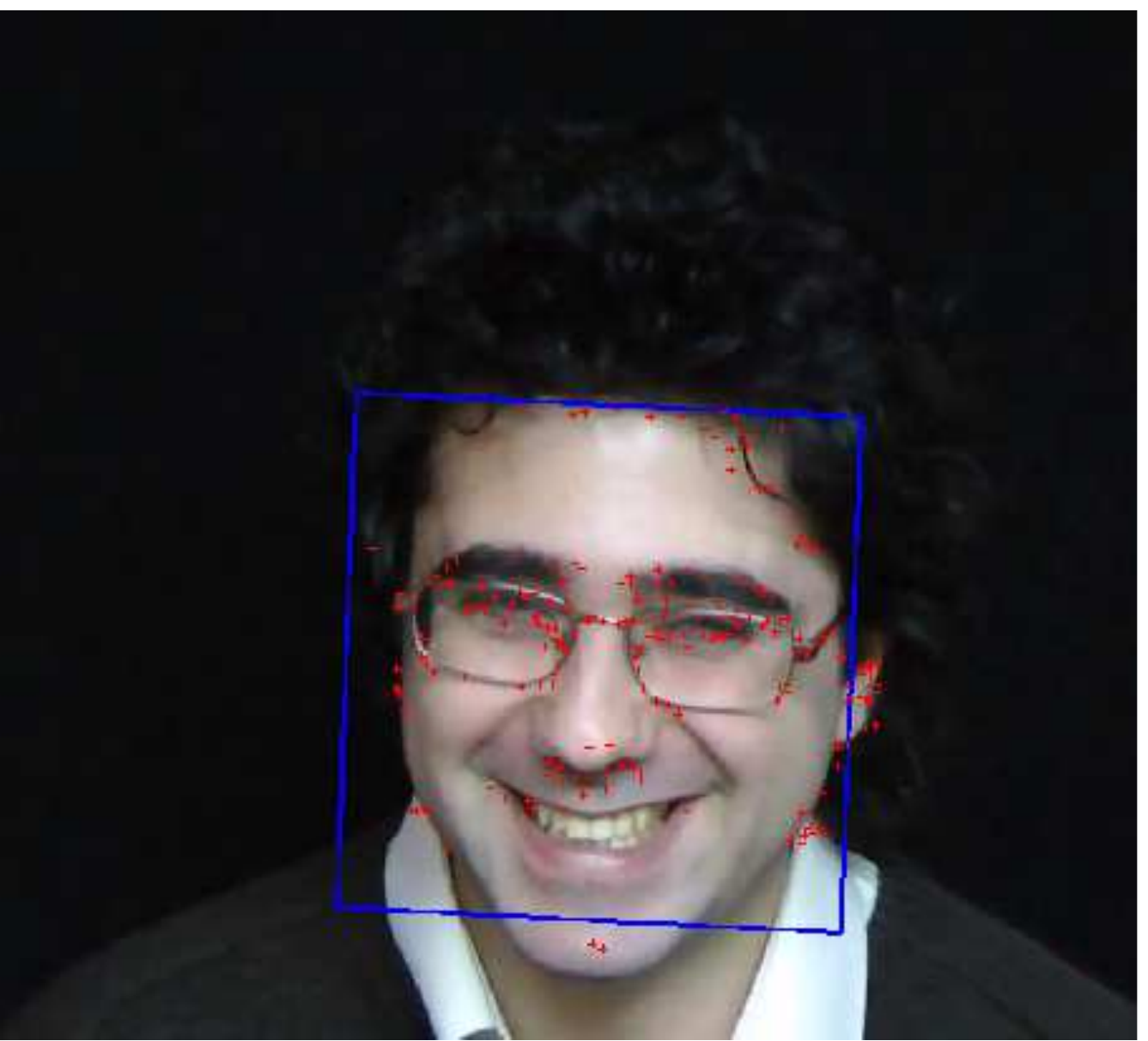}}}
\caption{KLT tracker in action.}\label{KLTtracker}
\end{center}
\end{figure}

A Kanade-Lucas-Tomasi (KLT) tracker \cite{Tomasi1} was implemented to track the face and eyes in the video. The tracker searches for good feature points to track within the face and eyes bounding boxes and tracks these points across the video. These points are marked as crosses (+) in the faces shown in Fig. \ref{KLTtracker}. An affine transformation is estimated based on the movement of the feature points from one frame to the next, and the bounding box is then warped according to the transformation to keep track of the face region. Fig. \ref{KLTtracker} shows the original bounding box in the first frame of the video, and the warped bounding box in a later frame. The output of this step is the coordinates of the bounding box vertices at every frame. \vspace{-0.5cm}

\vspace{-1.1cm}
\subsubsection{Cropping Strategies:}
The KLT tracker is used to track the face region in the video. Then, three different methods of normalization are tested.
\begin{enumerate}
\item \emph{Normalize eye location:} This method normalizes the location of the eyes in the video, as described by Mandal \emph{et al.} in \cite{Mandal5,Mandal8}. The location of the eyes are tracked, and the video is rotated and resized such that both eyes are always on the same horizontal axis, and there is a fixed pixel distance between both eyes (234 pixels apart for a face image of $400\times500$, similar to \cite{FERETnorm,Mandal2,Jiang5}). Fig. \ref{eyeLocNorm} shows a sample image of three subjects from UvA-NEMO smile database.
    \begin{figure}[!ht]
    \begin{center}
    \scalebox{0.2}{\rotatebox{0}{\includegraphics*{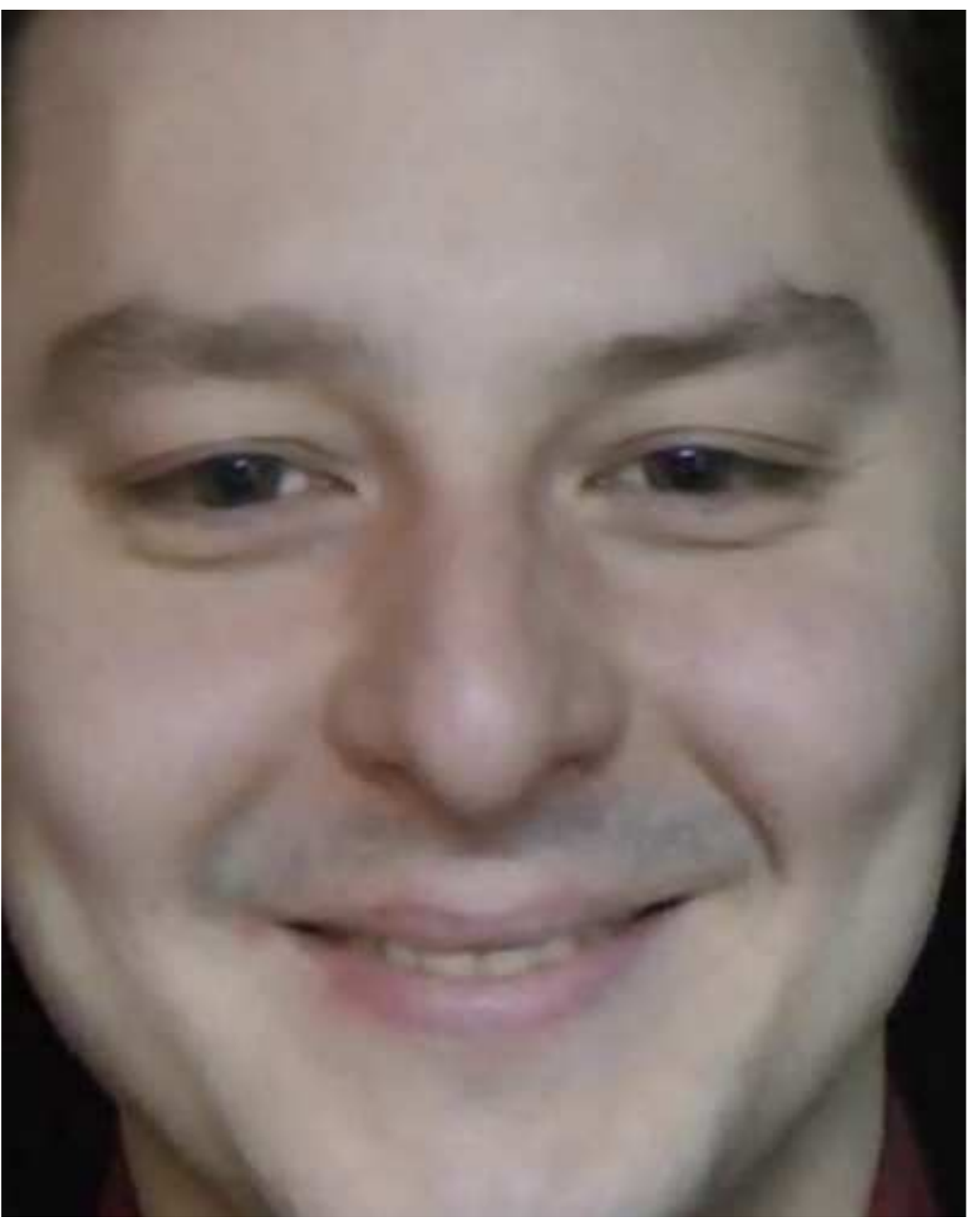}}}
    \scalebox{0.2}{\rotatebox{0}{\includegraphics*{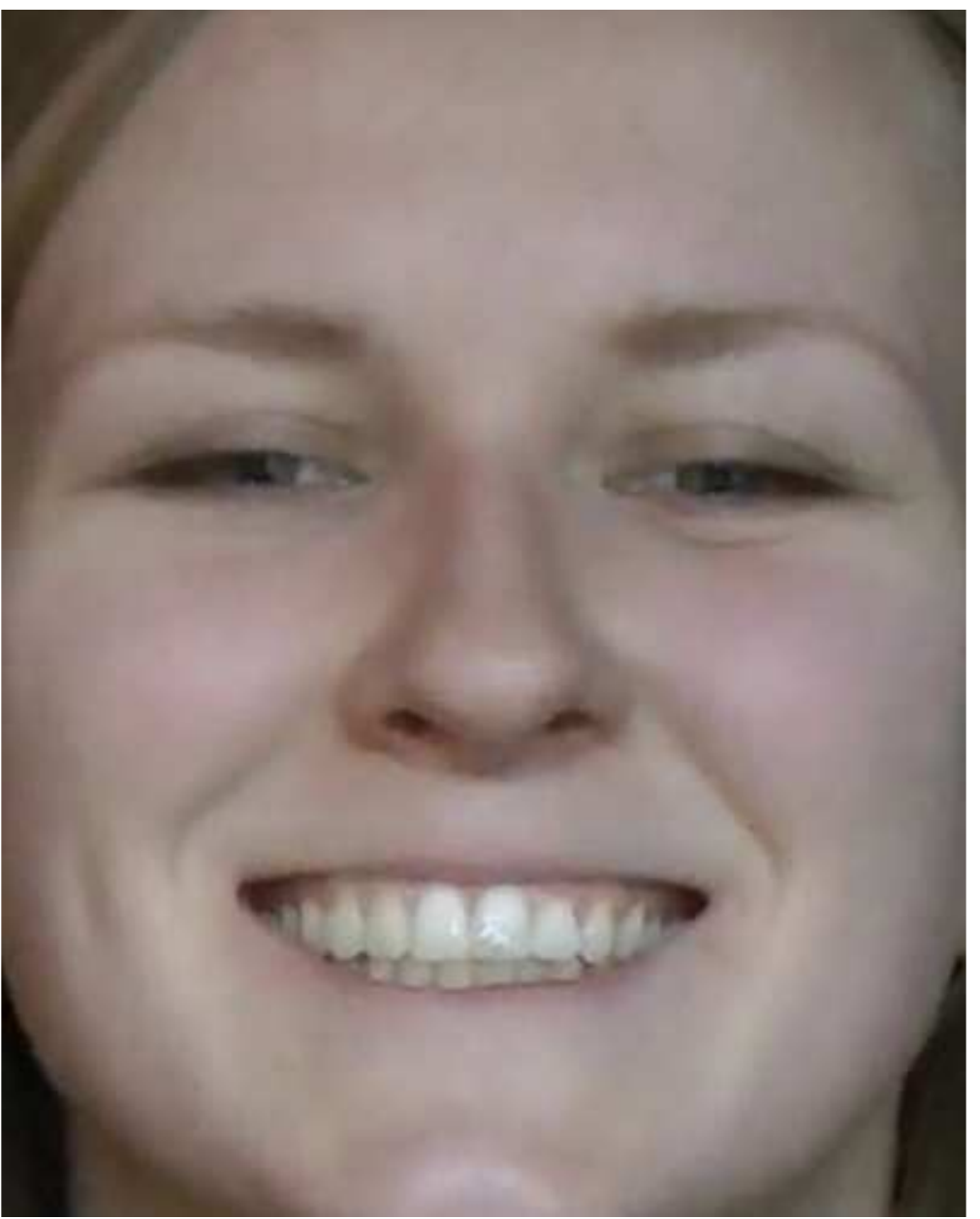}}}
    \scalebox{0.2}{\rotatebox{0}{\includegraphics*{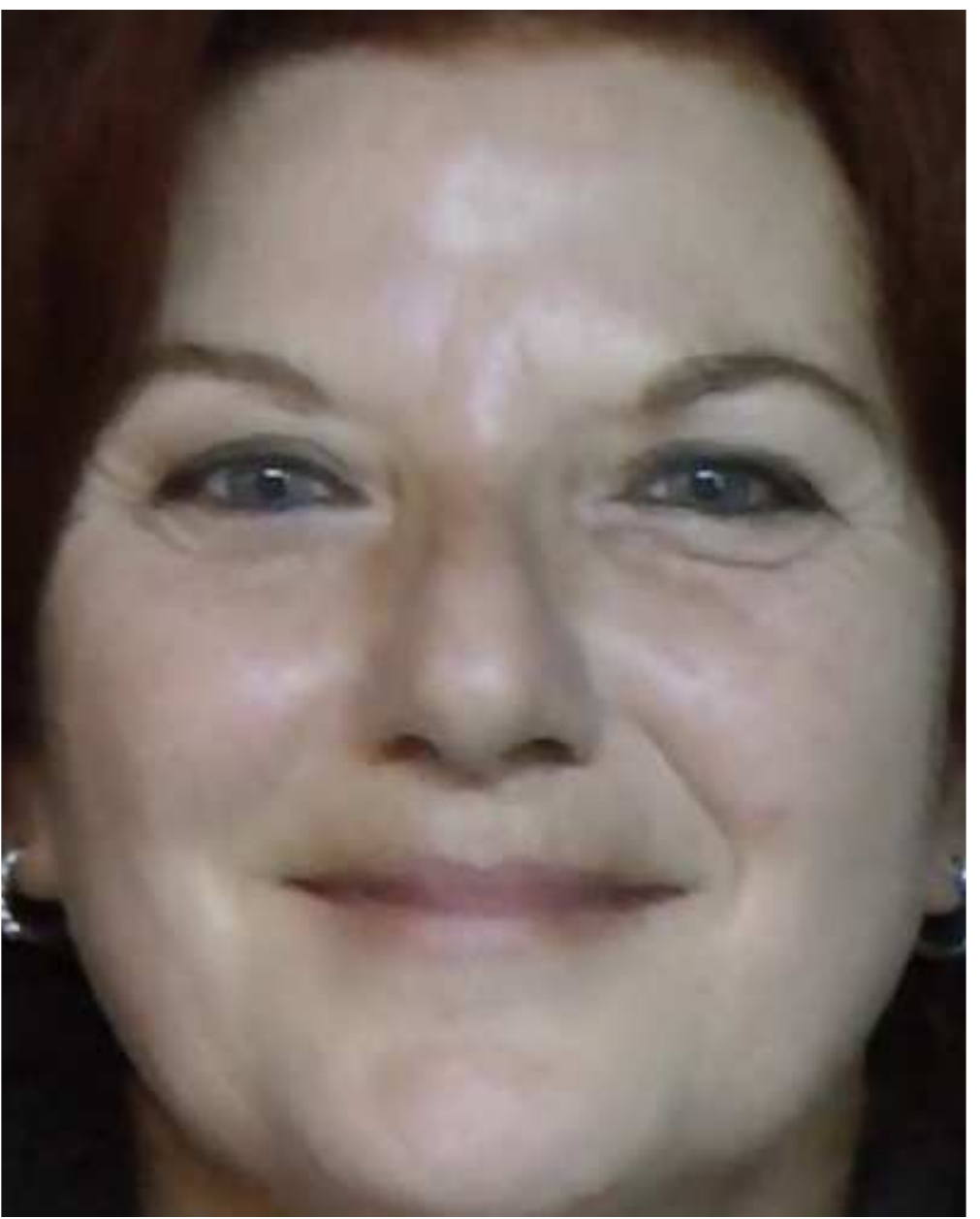}}}
    \caption{Cropped face regions after eye location normalization.}\label{eyeLocNorm}
    \end{center}
    \end{figure}
\item \emph{Normalize face orientation:} This method normalizes the orientation of the face. The orientation of the face is obtained from the KLT tracking data, and the video is rotated to compensate for the face rotation such that the face appears upright. Fig. \ref{faceLocNorm} shows the corresponding samples.
    \begin{figure}[!ht]
    \begin{center}
    \scalebox{0.2}{\rotatebox{0}{\includegraphics*{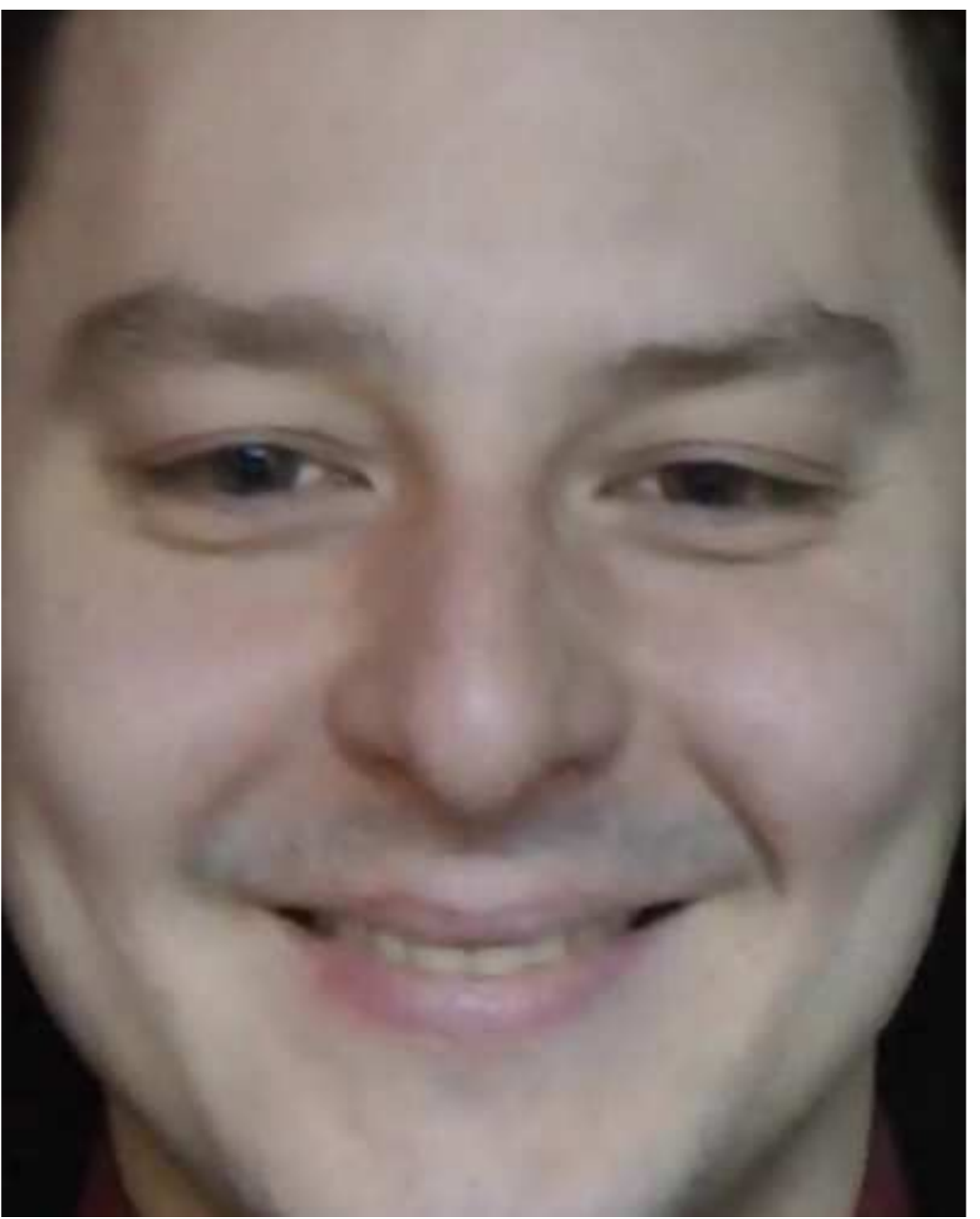}}}
    \scalebox{0.2}{\rotatebox{0}{\includegraphics*{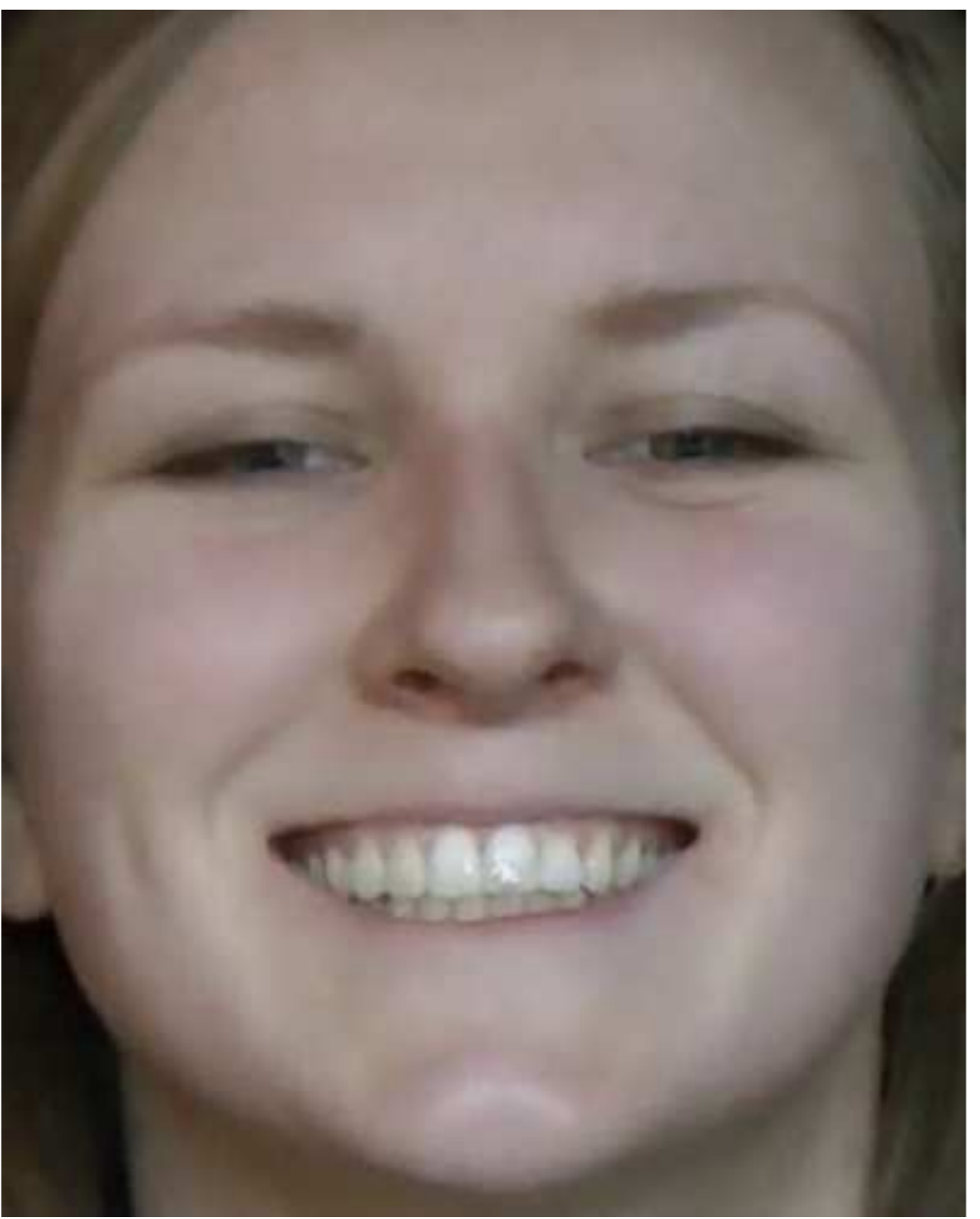}}}
    \scalebox{0.2}{\rotatebox{0}{\includegraphics*{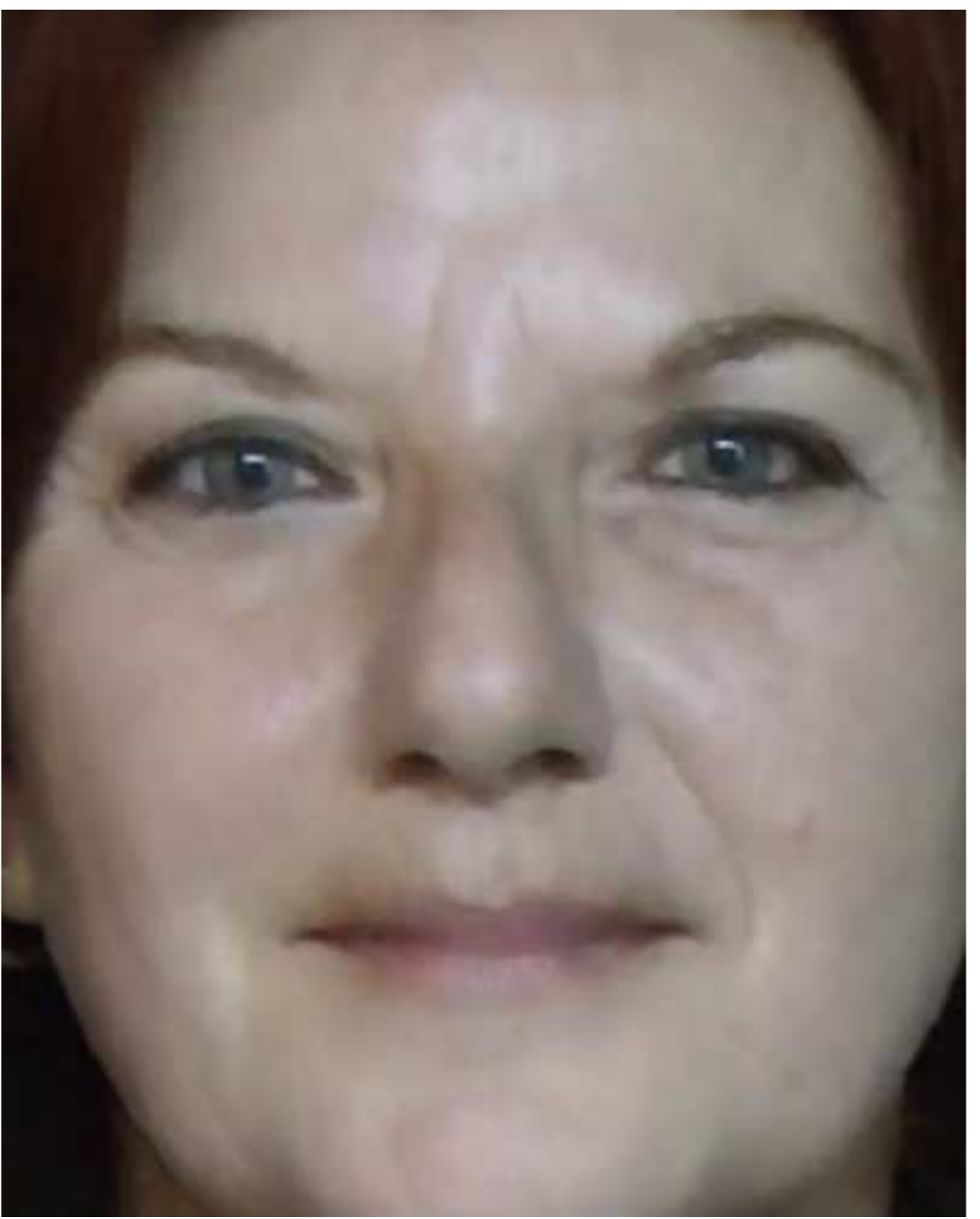}}}
    \caption{Cropped face regions after rotation normalization.}\label{faceLocNorm}
    \end{center}
    \end{figure}
\item \emph{No normalization:} This is a control experiment to observe the effects of normalization. Fig. \ref{noNorm} shows the corresponding samples.
    \begin{figure}[!ht]
    \begin{center}
    \scalebox{0.2}{\rotatebox{0}{\includegraphics*{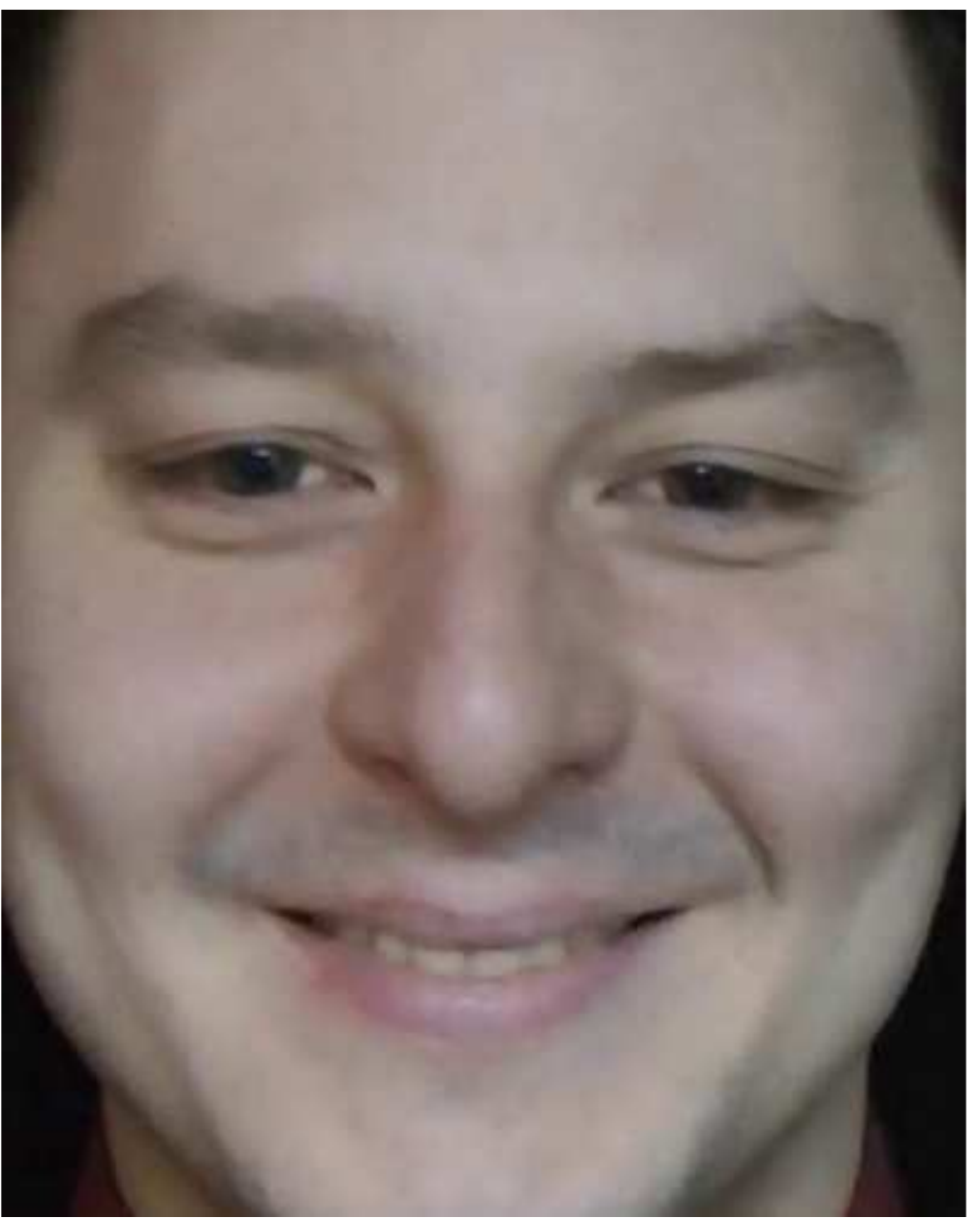}}}
    \scalebox{0.2}{\rotatebox{0}{\includegraphics*{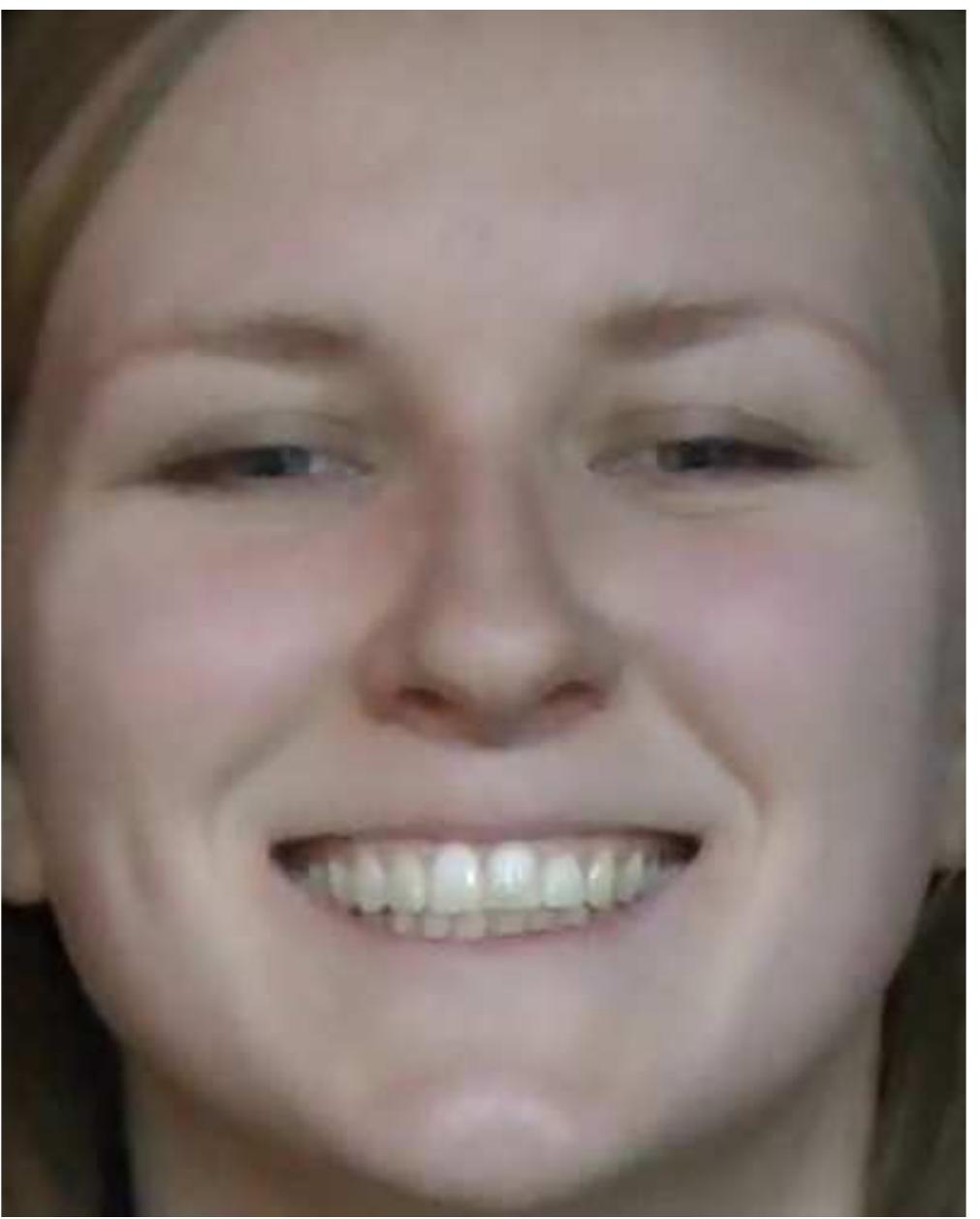}}}
    \scalebox{0.2}{\rotatebox{0}{\includegraphics*{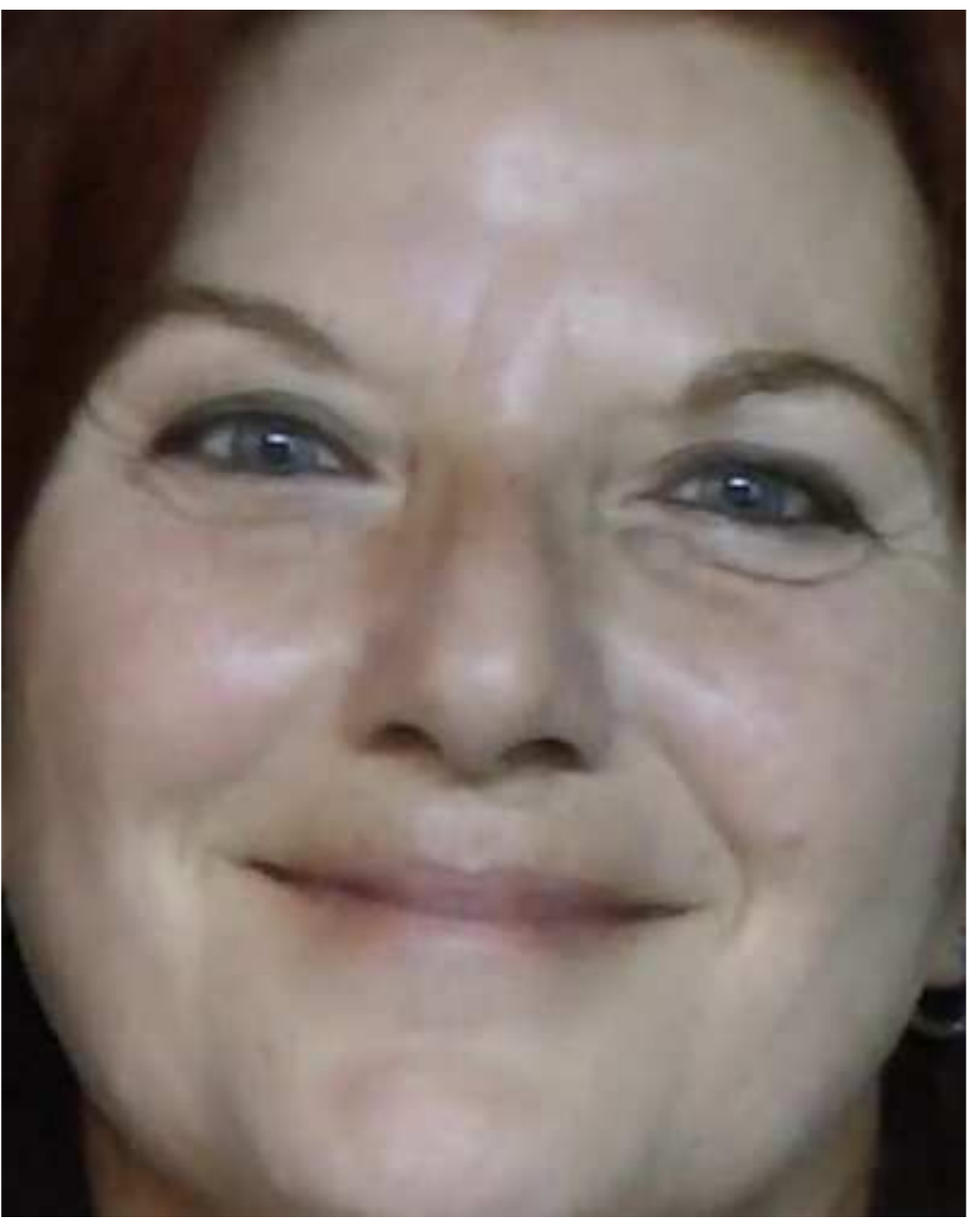}}}
    \caption{Cropped face regions with no normalization.}\label{noNorm}
    \end{center}
    \end{figure}
\end{enumerate}
After normalization from each of these techniques, the face region is cropped out of the video for feature extraction. The first two methods produce a $400\times500$ pixel region, while the third method produces a $720\times900$ pixel region.

\subsection{Micro-expression Amplification}
Eulerian Video Magnification (EVM) is a Eulerian method to amplify small motions and color variations in videos. It was first developed by Wu \emph{et al.} \cite{Wu1}, and the motion amplification was later improved with a phase-based pipeline by Wadhwa \emph{et al.} \cite{Wadhwa1}. It is used to amplify micro-expressions in the extracted face region. This method of motion amplification has shown good results when used in a system to detect spontaneous micro-expressions \cite{Li5}. The algorithm first decomposes an input video into different spatial frequency bands and orientations using complex steerable pyramids, and the amplitude of local wavelets are separated from their phase.  The phases at each location, orientation and scale are then independently filtered and the phase of a specified frequency range is amplified (or attenuated) and de-noised. Finally, the video is reconstructed, which now has some motions amplified, depending on the frequencies selected.

An original image $X$ is passed through the pre-trained VGG network and feature map $X^f$ of size $w\times h \times d$ is obtained. Where $w$ and $h$ are the width and height of the feature map obtained from $23^{rd}$ of the VGG face network, there are $d$ such feature maps. A gaussian mask $G$ is created with center as fixation point $x_a$ of image size $w\times h$ and variance 10. Each of these feature maps are multiplied with the Gaussian mask $G$, given by:
\begin{equation}
\widetilde{X_i}=X^f * G,
\end{equation}
All $\widetilde{X_i}$, where $i \in \{1,\ldots, d\}$ are converted into a column vector by lexicographic ordering the feature map elements of each of the $\widetilde{X_i}$, which is our data input vector $\mathbf{x}_i$.
A temporal sequence of vectors $(\mathbf{x}_1, \mathbf{x}_2, \ldots, \mathbf{x}_T)$ as inputs. The outputs are the high level representations, is a sequence of vectors $(\mathbf{c}_1, \mathbf{c}_2, \ldots, \mathbf{c}_T)$. These representations are obtained by non-linear transformation of the input sequences of vectors from $t=1$ to $T$. These are given by:

\begin{equation}
\mathbf{c}_t=f(W\mathbf{x}_t+J\mathbf{c}_{t-1}+\mathbf{b})
\end{equation}

\begin{equation}
\mathbf{p}_t=softmax(\mathbf{W}_y\mathbf{c}_t+\mathbf{b}_y)
\end{equation}

Where the non-linear function $f$ is applied element-wise, and $\mathbf{p}_t$ is the $softmax$ probabilities of the video frames having seen the observations up to $\mathbf{x}_t$. Matrices $\mathbf{W}$, $\mathbf{J}$, $\mathbf{b}$, $\mathbf{W}_y$ and $\mathbf{b}_y$ are the parameters learned during training. We apply the function $f$ as $ReLU$, which is given by:

\begin{equation}
f(x)=ln(1+e^x)
\end{equation}

\begin{equation}
p(x_a)=exp(-\frac{\|x_a - x_i\|^2}{(s\sigma_i)^2})
\end{equation}

\begin{equation}
p(x_a)=exp(-\frac{\|x_a - x_i\|^2}{(s\sigma_i)^2})
\end{equation}

\subsection{Feature Extraction}
In this subsection details of the feature extraction and their corresponding normalization techniques are discussed. Features are extracted from the cropped face region. Four different image-processing techniques are used to extract features.
\begin{enumerate}
\item \emph{Local Phase Quantization:} LPQ was originally used for texture description \cite{Ojansivu1}, which outperformed the Local Binary Pattern operator in texture classification and has been extended to face recognition \cite{Ahonen2}. LPQ features are insensitive to motion blurs, out of focus blurs, and atmospheric turbulence blurs. The algorithm computes a 2D Discrete Fourier Transform (DFT) over a neighborhood at each pixel using 2D convolutions. Then, the covariance matrix of the DFT is obtained, de-correlated and quantized to obtain the local phase information. LPQ descriptors are less sensitive to image blurs which may arise from the interpolation of the normalization techniques, and can be useful for smile recognition.
    \\
\item \emph{Histogram-of-Oriented-Gradients:} HOG is typically used for objection detection, and has been used for human detection in videos \cite{Dalal1}. The idea behind HOG is that an object's shape can be well-characterized by the distribution of local intensity gradients (edge directions) without precise knowledge of their positions. The algorithm divides the image into spatial regions called cells, and a histogram of gradient directions is computed for each pixel in the cell. Cells are grouped to form larger spatial regions called blocks, and the local histogram `energy' of a block is used to normalize the contrast in all the cells in that block. These normalized blocks are the HOG descriptors. The `UoCTTI' variant of HOG \cite{Felzenszwalb1} is used as it has a 31 features per cell compared to the original 36 features per cell, which is a significant reduction in dimensionality. A $4\times4$ pixel cell with 8 orientations is used as early experiments show that it has the best results for the given problem.
    \\
\item \emph{Dense Optical Flow:} Optical flow is the pattern of apparent motion of objects in a visual scene caused by a relative movement between an observer and the scene. Optical flow can be computed for a sparse feature set to track the movement of the feature points, or for a dense feature set whereby every pixel in an image sequence is tracked. Dense optical flow is used to calculate the movement of each pixel throughout the video. This work uses a dense optical flow field determined by a differential method proposed by Farneback \cite{Farneback1}. Optical flow works on the assumptions that the pixel intensities of a scene do not change between consecutive frames, and that neighboring pixels have similar motion. Farneback's algorithm approximates an image signal to be a polynomial expansion and solves for the displacement of that polynomial assuming a translation occurred. The solution is made to be more robust by introducing assumptions to constrain the system. The main assumptions made are that the displacement field is slowly varying, and that the displacement field can be parameterized according to a motion model.
    \\
\item \emph{Pre-trained Convolutional Neural Network (CNN) Features:} Another emerging class of features are the pre-trained deep CNN features. Studies have shown that deep convolutional neural network (CNN) model trained for certain application can be applied to similar classification problems perform well \cite{Hu1,Parkhi1}. We use the pre-trained deep CNN model for face recognition, which is trained with 2.6 million images comprising of over 2600 people for face smile classification. Most existing models related to face or object recognition is applied to images, not videos. To accommodate this method for video-processing, each frame of the video is processed and features are extracted from a deep fully-connected layer. These features are post-processed to combine them and used for classification with a SVM. This work uses the VGG face recognition model \cite{Parkhi1}, and 4096 features with L2 normalized per frame are extracted from the $35^{th}$ layer of the CNN network.
\end{enumerate}

The techniques mentioned above extract features frame-by-frame. Since the video samples have a varying number of frames, there is also a varying number of features per sample. Our learning algorithms require a fixed number of features per sample, so the number of features per sample is normalized. To do so, the features of each video are concatenated across time and transformed to the cosine domain via Discrete Cosine Transform, such that the number of cosine coefficients is equal to the number of frames in the video. Then, the number of coefficients are normalized by either cropping or padding them such that there is an equal number of cosine coefficients per sample video; this is equivalent to normalizing the number of frames per video. The number of frames in the videos range between 80-500 frames, and the average number of frames is 150. A suitable number of frames for normalization would be 150, however computational constraints limit the number of features that each sample can have, thus the number of frames per video is normalized to 100 for the experiments.

\subsection{Classification}
A SVM is a machine learning algorithm that can be used for classification, regression or other tasks by constructing a hyperplane or a set of hyperplanes. The modern implementation of SVMs was developed by Cortes and Vapnik in \cite{Cortes1}. Recent study shows that when applied to distinguishing posed and spontaneous smiles, the linear SVM classifier outperforms other classifiers like the Polynomial Kernel SVM, RBF Kernel SVM, Linear Discriminant, Logistic Regression, k-Nearest Neighbour and Naïve Bayes \cite{Dibeklioglu3}. Therefore, linear SVM is used for classification in this work.

\section{Experiments}
\label{sec:blind}

The UvA-NEMO Smile Database is used to analyze the dynamics of posed and spontaneous smiles \cite{Dibeklioglu1}. This is the largest database of videos of spontaneous and posed smiles to date \cite{Mandal11}, with 643 posed smiles and 597 spontaneous smiles from 400 subjects. The videos were recorded with a Panasonic HDC-HS700 3MOS camcorder that was placed on a monitor approximately 1.5m away from the recorded subjects. The videos have a resolution of $1920\times1080$ pixels and a frame rate of 50 frames per second. For posed smiles, each subject was asked to pose an enjoyment smile after being shown a demonstration video. For spontaneous smiles, each subject was shown a set of short, funny videos for approximately 5 minutes. Two trained annotators segmented the recorded video of their smiles to obtain the smile video.

Similar to \cite{Dibeklioglu1,Mandal11}, the accuracy of the system is measured using 10-fold cross validation, where the data set is divided into 10 folds specified by the smile database. 9 folds are used for training and the remaining fold is used for testing. This is done 10 times, such that each fold is used for testing once in order to represent the entire data set, and the average of the results is used as the measure of accuracy. The classification accuracy is recorded into a confusion matrix. To simplify the presentation of results, only the accuracy of true positives and negatives are displayed. The large number of features from HOG coupled with the larger face region size from not normalizing the face results in a large dimensionality that could not be computed using the SVM classifier. Therefore, the results for that case is omitted.

\begin{figure}
\begin{subfigure}{.6\textwidth}
  \centering
  \includegraphics[width=.3\linewidth]{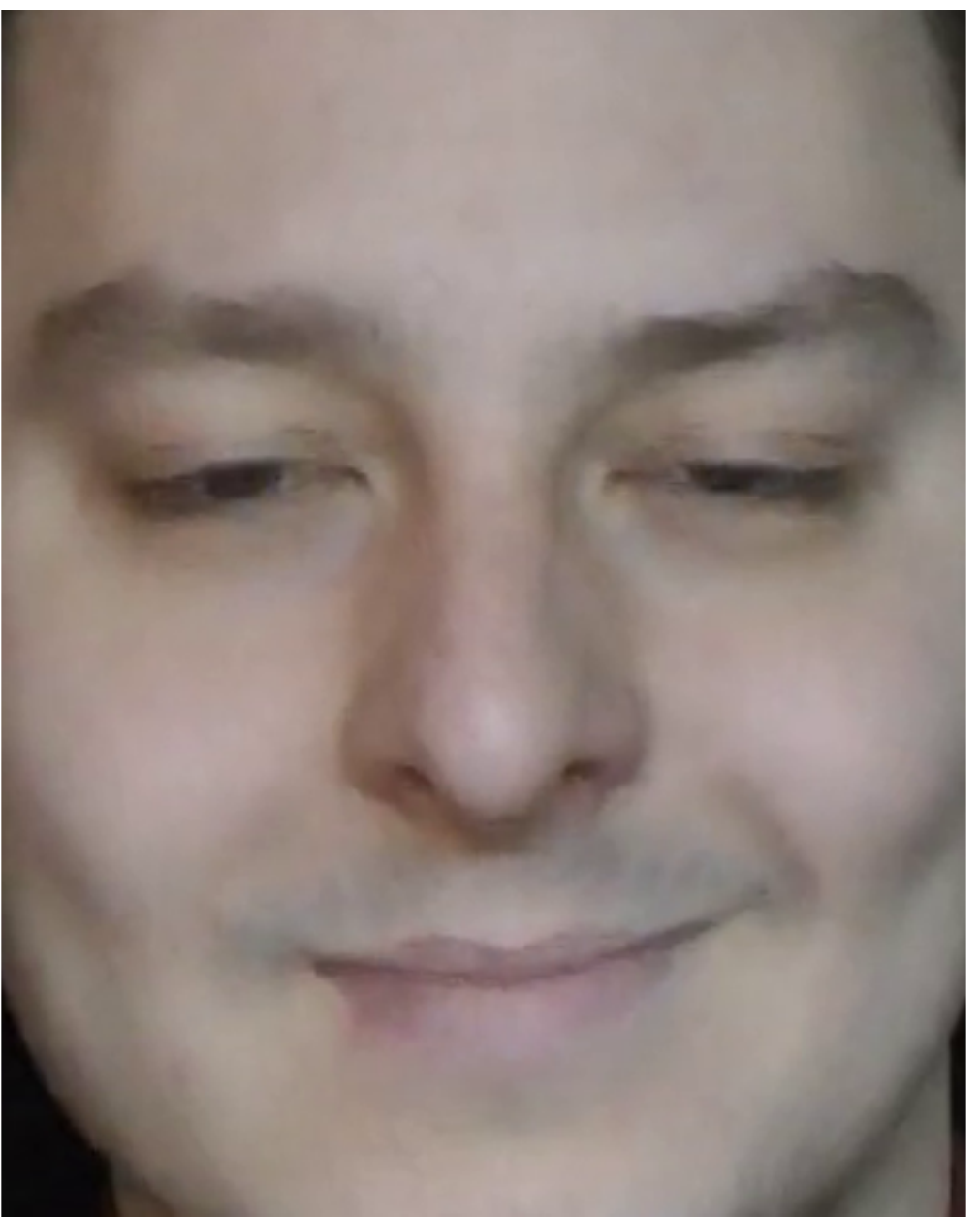}
  \includegraphics[width=.3\linewidth]{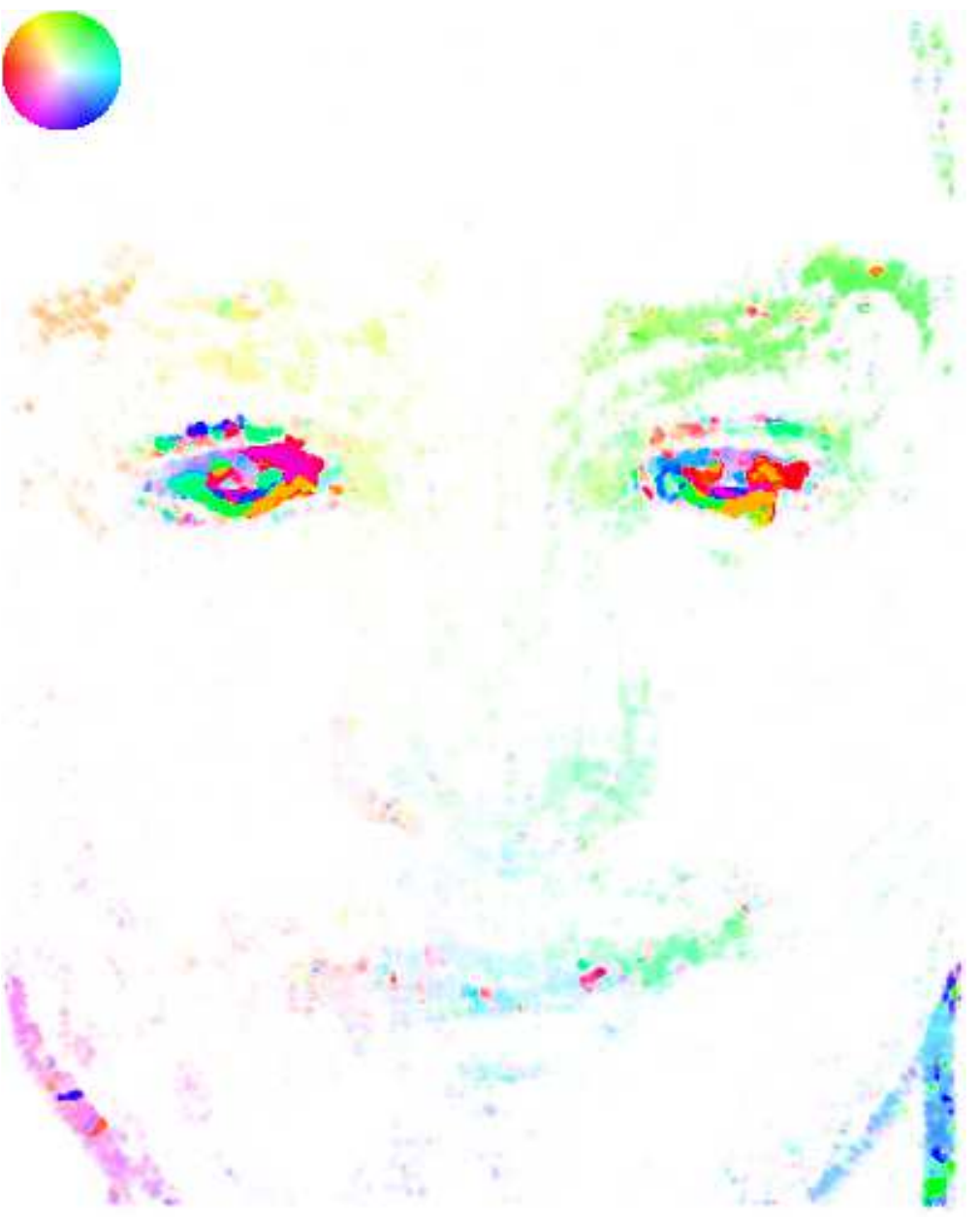}
  \caption{Optical flow of a subject blinking while smiling.}
  \label{fig:sfig1}
\end{subfigure}%
\begin{subfigure}{.3\textwidth}
  \centering
  \includegraphics[width=.6\linewidth]{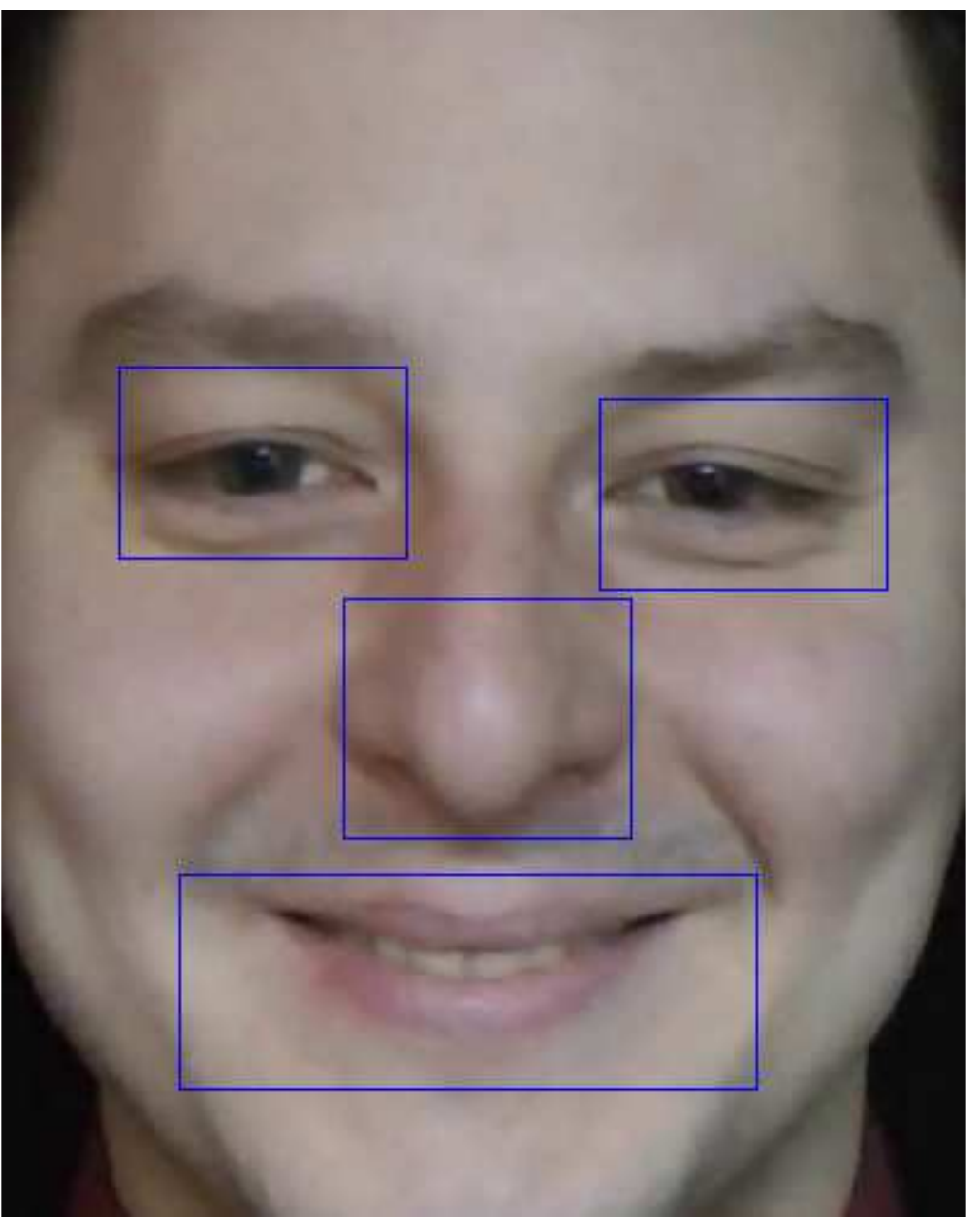}
  \caption{Region extraction.}
  \label{fig:sfig2}
\end{subfigure}
\caption{Left two: optical flow of a blink, Right: extracted regions}
\label{fig:fig}
\end{figure}

Fig \ref{fig:sfig1} shows the optical flow output, where the hue of the pixel denotes the flow direction, while the saturation of the pixel denotes the flow magnitude. The drawback of dense optical flow is that a large number of features is produced. For a face region of $400\times500$ pixels, each pixel will have 2 features for optical flow in the x- and y-directions. This results in 400,000 features per frame, and approximately 30 million features per video. The dimensionality for classification is too high, requiring a very long time and a large memory to process the training data. Therefore, the dimensionality must be reduced in order for this method to be feasible. To do so, the optical flow data from the eyes, nose and mouth regions are extracted and used for classification, while the other regions are discarded since they are not likely to have a large impact on classification. Fig. \ref{fig:sfig2} shows the bounding boxes of the regions that are extracted.

The eye regions are obtained from the tracking data. The nose and mouths regions are estimated based on the location of the eyes. The result is a dimensionality of approximately 52,800 features per 21 frame, and approximately 8 million features per video. However, the accuracy obtained using 10-fold cross validation (see Sections 3 and 4) for dense optical flow is bad, with 59.9\% accuracy for classifying posed smiles correctly, and 70.3\% accuracy for classifying spontaneous smiles. Due to the long time that it takes to extract dense optical flow features and the large dimensionality of the features, it was deemed that this method is not worth further tests and is omitted from the discussions hereafter.

Table \ref{Tab_1} shows the classification accuracy of the system by varying the normalization methods, feature extraction techniques, and the use of EVM for micro-expression amplification. For example, the top-left entry of the table shows that there is a 81.7\% accuracy for classifying posed smiles, and 74.3\% accuracy for classifying spontaneous smiles, when the eye location is normalized, EVM is used to amplify micro-expressions, and HOG features are used. The table abbreviates the pre-trained VGG convolutional neural network model features to `VGG'. From Table \ref{Tab_1}, it can be seen that the classification of posed smiles is always more accurate than spontaneous smiles, regardless of normalization or feature extraction method. This may be a result of having more prominent visual appearances of the posed smiles as compared to the spontaneous ones. Perhaps, this allows the SVM to have better support vectors to define posed smiles, whereas lesser visual appearances of spontaneous smiles could mean that the support vectors are less well defined and the hyperplane produced is not a good true separator of boundaries.
\begin{table}[!htp]
\caption{True positive (posed smiles) and true negative (spontaneous smiles) classification accuracy (\%) of our system with varying feature extraction methodologies and normalization.} \label{Tab_1} \centering
\begin{tabular}{|c|c|c|c|c|c|c|}
\hline
 & \multicolumn{2}{|c|}{Norm. Eye Location} &\multicolumn{2}{|c|}{Norm. Face Orientation} &\multicolumn{2}{|c|}{No Normalization}\\
\hline
& EVM & No EVM & EVM & No EVM & EVM & No EVM\\
\hline
HOG & $81.7, 74.3$ & $82.8, 72.6$ & $82.4, 71.6$ & $81.7, 71.7$ & - & -\\
\hline
LPQ & $83.9, 65.1$ & $80.1, 70.8$ & $84.7, 65.1$ & $79.9, 70.8$ & $84.3, 65.3$ & $82.6, 66.0$\\
\hline
VGG & $82.9, 70.6$ & $82.9, 70.3$ & $82.7, 69.7$ & $82.2, 71.6$ & $82.5, 69.6$ & $82.4, 69.3$\\
\hline
\end{tabular}
\end{table}

\begin{table}[!htp]
\caption{Overall classification accuracy (\%) of our system with varying feature extraction methodologies and normalization.} \label{Tab_2} \centering
\begin{tabular}{|c|c|c|c|c|c|c|}
\hline
 & \multicolumn{2}{|c|}{Norm. Eye Location} &\multicolumn{2}{|c|}{Norm. Face Orientation} &\multicolumn{2}{|c|}{No Normalization}\\
\hline
& EVM & No EVM & EVM & No EVM & EVM & No EVM\\
\hline
HOG & $78.14$ & $77.89$ & $77.20$ & $76.89$ & - & -\\
\hline
LPQ & $74.85$ & $75.62$ & $75.26$ & $75.52$ & $75.15$ & $74.61$\\
\hline
VGG & $76.98$ & $76.83$ & $76.44$ & $77.10$ & $76.29$ & $76.09$\\
\hline
\end{tabular}
\end{table}
Table \ref{Tab_2} shows that overall accuracy of our proposed approaches. Although VGG face model is trained with millions of face images, features from such deep CNN network may not be good for face smile classification. It is evident that HOG features with normalization using the eye locations along with the magnified micro-expression using EVM performs best in most of the cases as compared to LPQ and VGG features.

\subsection{Comparison with Other Methods}
Correct classification rates (\%) using various methods on UvA-NEMO are shown in Table \ref{CCRComparison}. Most of the existing methedologies involve semi-automatic processes for face feature extraction, whereas our method is fully automatic and does not require any manual intervention. It is evident from the table that our proposed approach is quite competitive as compared to the other state-of-the-arts methodologies. \vspace{-0.7cm}
\begin{table}
\caption{Correct classification rates (\%) on UvA-NEMO database.}
\footnotesize
\centering
\begin{tabular}{|c|c|*{2}{c|}}
\hline
Method & Process & Correct Classification Rate (\%)\\
\hline
Pfister \emph{et al.} \cite{Pfister1} & semi-automatic & 73.1\\
\hline
Dibeklioglu \emph{et al.} \cite{Dibeklioglu2} & semi-automatic & 71.1\\
\hline
Cohn \& Schmidt \cite{Schmidt1} & semi-automatic & 77.3\\
\hline
Eyelid Features \cite{Dibeklioglu1} & semi-automatic & 85.7\\
\hline
Eye+Lips+dense optical flow \cite{Mandal11} & semi-automatic & 80.4\\
\hline
\textit{Our proposed method} & fully automatic & 78.1\\
\hline
\end{tabular}
\label{CCRComparison}
\end{table}

\vspace{-0.7cm}
\subsection{Discussions}
\begin{figure}[!ht]
\begin{center}
\scalebox{0.55}{\rotatebox{0}{\includegraphics*{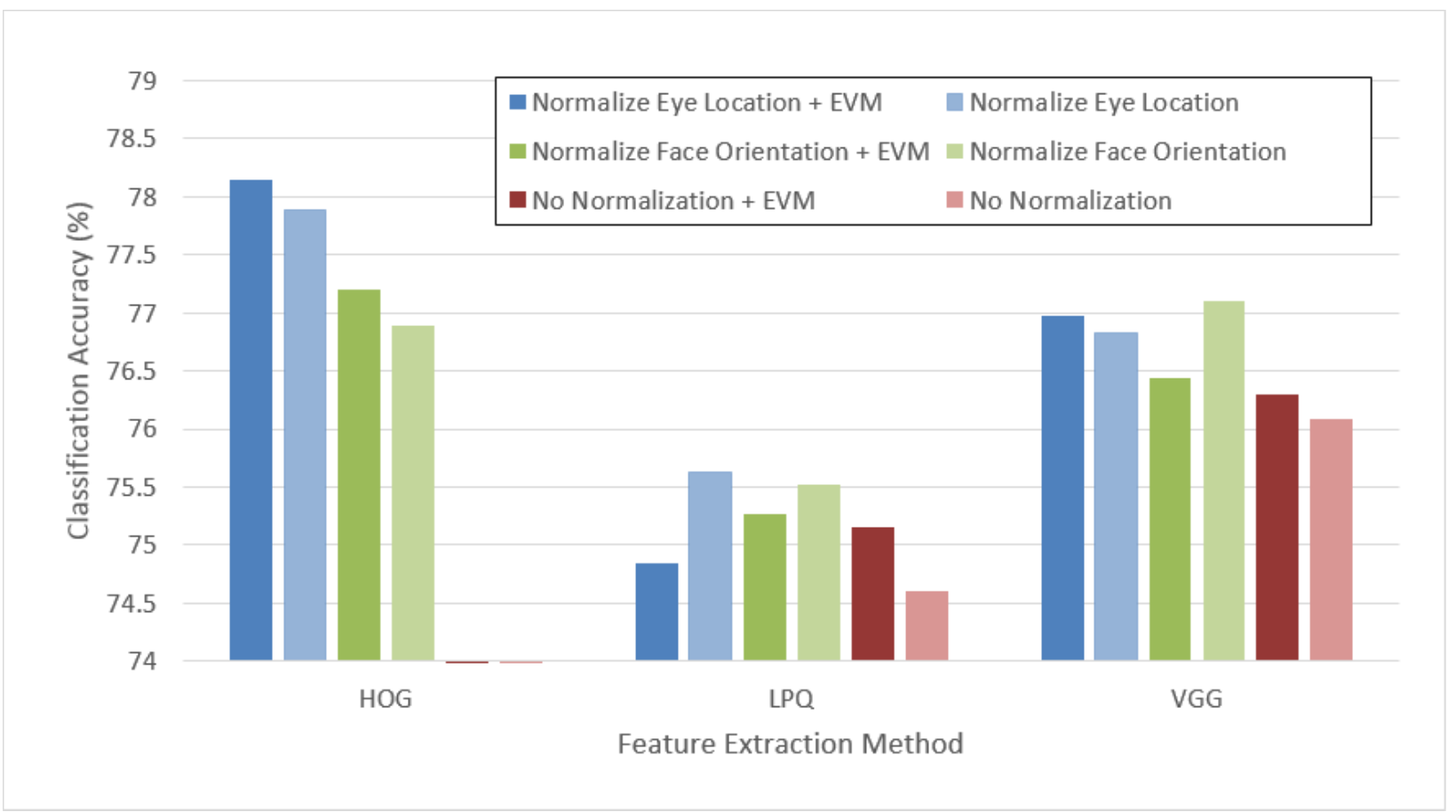}}}
\caption{Comparison of accuracy with feature extraction methods.}\label{featureExt}
\end{center}
\end{figure}

Table \ref{Tab_1} shows that the classification of posed smiles is always more accurate than spontaneous smiles, regardless of normalization or feature extraction method. This is probably because posed smiles may have more prominent visual appearances as compared to the spontaneous ones. Fig. \ref{featureExt} represents the results in Table \ref{Tab_2} in a histogram by grouping the feature extraction methods together. It can be seen that HOG outperforms the other two feature extraction methods, LPQ and VGG, for most of the normalization techniques, except for one case where using HOG has a 0.21\% lesser accuracy than using the pre-trained NN model features when normalizing the face orientation, and not using EVM. Similarly, using the pre-trained VGG CNN model outperforms LPQ as a feature extraction method as the classification accuracy using VGG is higher than LPQ for all cases.

From the experiments, it can be said that among the 3 feature extraction methods, HOG features are the best, and LPQ features are the worst. Features from pre-trained VGG model using large number of face images for face recognition has not been able to generalize well for smile classification. It is interesting to note that HOG is capable of capturing the fine grained facial features that helps in distinguishing posed and spontaneous smiles.
\begin{figure}[!ht]
\begin{center}
\scalebox{0.55}{\rotatebox{0}{\includegraphics*{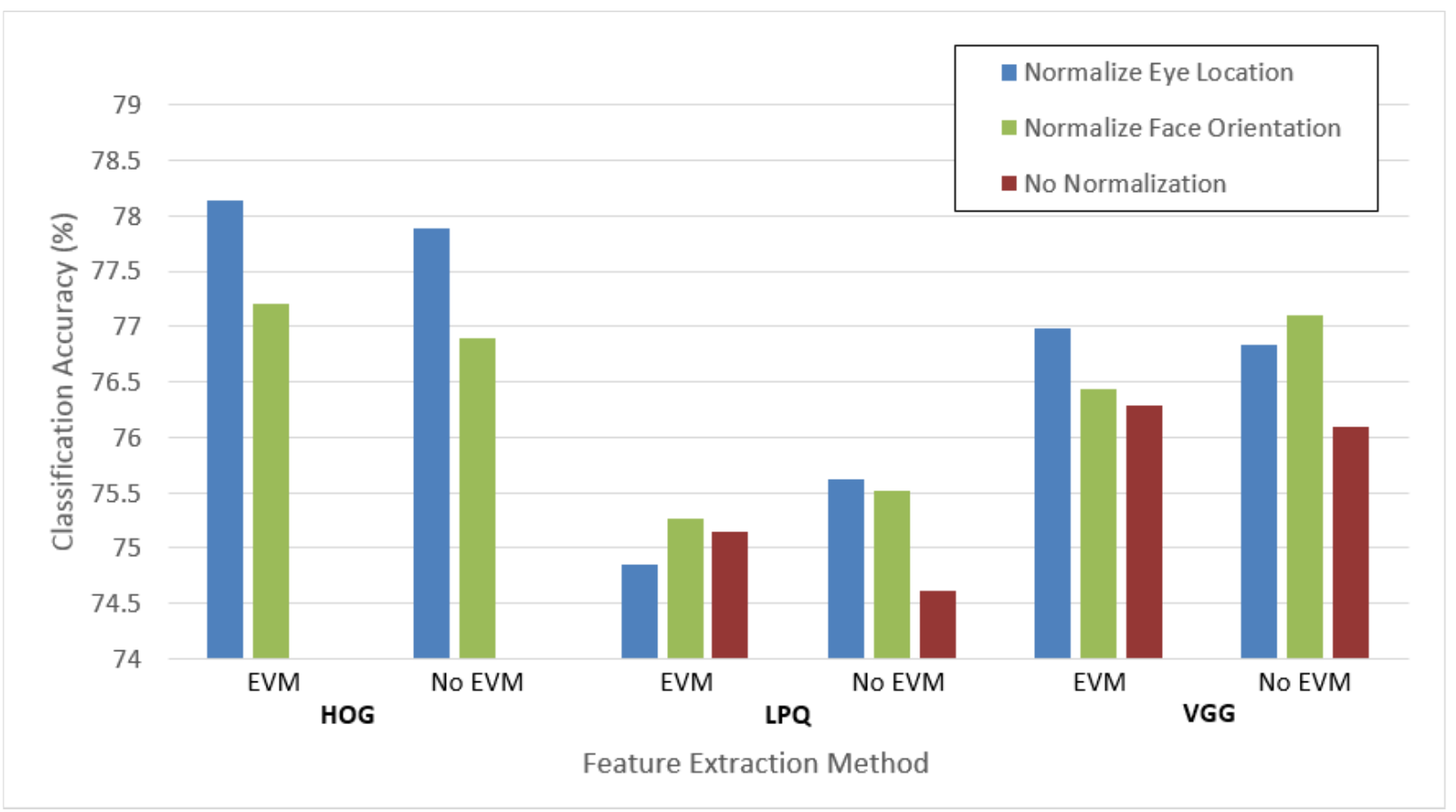}}}
\caption{Comparison of classification accuracy with normalization methods.}\label{classiNorm}
\end{center}
\end{figure}

Fig. \ref{classiNorm} shows the comparison of classification accuracy between normalization methods, which are grouped according to the parameters. Generally, having no face normalization results in a lower classification accuracy as compared to having either normalization method. The only exception is the case of using LPQ features with EVM, where the classification accuracy is 0.3\% higher without normalization than normalizing the eye location. Normalizing the eye location performs the best in 4 out of the 6 cases, thus it seems like it is the better normalization. The most significant difference between normalization techniques is seen when using HOG descriptors, where there is $~1$\% difference in classification accuracy between normalizing eye location vs. normalizing face orientation.

\section{Conclusions}
In this work, a cluster of methodologies is proposed to distinguish posed and spontaneous smiles. It involves four feature extraction methods, three normalization methods and the use of EVM for micro-expression amplification, but was unable to improve the state-of-the-art performance. The best classification accuracy obtained was 78.14\%. Using EVM to amplify micro-expressions did not have a significant impact on classification accuracy, while the normalizing facial features improved classification accuracy. The advantage of our proposed approaches as compared to other methods is that they are fully automatic. The effectiveness of the feature extraction methods in smile classification is ranked from most to least effective as follows: HOG, a pre-trained VGG CNN model features, LPQ and dense optical flow. This work succeeded in identifying techniques which are helpful and detrimental to smile classification. Experimental results on large UvA-NEMO smile database show promising results as compared to other relevant methods.

\bibliographystyle{splncs}
\bibliography{BiblioFeb2015}

\end{document}